%% file: main.tex
\documentclass[sigconf]{acmart}

\usepackage{color,graphicx,url,booktabs,color,balance}
\usepackage{CJKutf8}
\usepackage{multirow}
\renewcommand\footnotetextcopyrightpermission[1]{}

\begin{document}
\title{Finding Similar Medical Questions from \\ Question Answering Websites}

\author{Yaliang Li$^1$, Liuyi Yao$^2$, Nan Du$^1$, Jing Gao$^2$, \\ Qi Li$^3$, Chuishi Meng$^4$, Chenwei Zhang$^5$, and Wei Fan$^1$}
\affiliation{$^1$Tencent Medical AI Lab, Palo Alto, CA USA}
\affiliation{$^2$State University of New York at Buffalo, Buffalo, NY USA}
\affiliation{$^3$University of Illinois at Urbana-Champaign, Illinois, IL USA}
\affiliation{$^4$JD Urban Computing Business Unit, Beijing, China}
\affiliation{$^5$University of Illinois at Chicago, Chicago, IL USA}
\affiliation{$^1$\{yaliangli, ndu, davidwfan\}@tencent.com, $^2$\{liuyiyao, jing\}@buffalo.edu}
\affiliation{$^3$qili5@illinois.edu, $^4$meng.chuishi@jd.com, $^5$czhang99@uic.edu}

\begin{abstract}
The past few years have witnessed the flourishing of crowdsourced medical question answering (Q\&A) websites. Patients who have medical information demands tend to post questions about their health conditions on these crowdsourced Q\&A websites and get answers from other users.
However, we observe that a large portion of new medical questions cannot be answered in time or receive only few answers from these websites.
On the other hand, we notice that solved questions have great potential to solve this challenge.
Motivated by these, we propose an end-to-end system that can automatically find similar questions for unsolved medical questions. 
By learning the vector presentation of unsolved questions and their candidate similar questions, the proposed system outputs similar questions according to the similarity between vector representations. Through the vector representation, the similar questions are found at the question level, and the diversity of medical questions expression issue can be addressed. Further, we handle two more important issues, i.e., training data generation issue and efficiency issue, associated with the LSTM training procedure and the retrieval of candidate similar questions. 
The effectiveness of the proposed system is validated on a large-scale real-world dataset collected from a crowdsourced maternal-infant Q\&A website. 
 \end{abstract}

\maketitle

\input{subfile/1_introduction}

\input{subfile/2_methodology}

\input{subfile/3_exp}

\input{subfile/4_related}
\input{subfile/5_conclusion}

\bibliographystyle{ACM-Reference-Format}
\balance

\end{document}

%% file: subfile/1_introduction.tex
\section{Introduction}
\label{sec:intro}

Nowadays, online medical services have become a rapidly growing industry. 
Researchers point out that the market size of online medical service in China reaches RMB $11.15$ billion with a growth rate of 128\% in $2016$\footnote{\url{https://bg.qianzhan.com/report/detail/459/171110-66a2d397.html}}.
It is hard for traditional offline medical services to meet the increasing needs of patients and this situation can be alleviated through online medical services. With the boom of the Internet and smartphones, people tend to get medical services, such as health information query, disease diagnosis, and health condition monitoring via various online services. For example, statistics indicate that more than $26$ million people search health-related information on Baidu search engine every day\footnote{\url{http://www.ebrun.com/20150812/144515.shtml}}.

The tremendous demand for online health services motivates the flourishing of crowdsourced question answering (Q\&A) websites. Patients who post questions on the crowdsourced Q\&A websites can get answers and feedback from other website users. Because of this feature, medical crowdsourced Q\&A websites attract increasingly more users. For example, one of the maternal-infant Q\&A website in China, \textit{baobaozhidao}\footnote{\url{baobao.baidu.com}}, has more than $21$ million registered users. Another medical crowdsourced Q\&A website in China, \textit{chunyuyisheng}\footnote{\url{www.chunyuyisheng.com}}, has more than $92$ million registered users. These crowdsourced Q\&A websites have dramatically changed the way that people seek health and medical information.

However, the growing number of users and posted questions on various crowdsourced Q\&A websites brings one major challenge: A large portion of new questions cannot be answered in time.
For example, there are averagely more than two thousand health and medical related new questions posted per hour on \emph{Baidu Knows}\footnote{https://zhidao.baidu.com}, the biggest online crowdsourced Q\&A website in China. While, most of them are not answered in time, even worse, many of these questions are not answered by any website user, because the amount of newly generated questions goes far beyond the capacity of answer providers.

Meanwhile, the treasure of solved questions is not fully explored, and such treasure is helpful to solve the aforementioned challenge. Crowdsourced Q\&A websites have a great number of already solved questions. For example, \textit{baobaozhidao} has nearly $55$ million maternal-infant related Q\&A pairs; \textit{chunyuyisheng} has more than $95$ million medical related Q\&A pairs. For a newly posted medical question, it is likely that there exist some semantically similar questions that are already solved. Thus the key to tackling the aforementioned challenge is to efficiently find questions from the large pool of solved questions that are similar to the newly posted questions. By this means, answers to the new questions can be capitalized on the knowledge from solved question-answer pairs. At the same time, the user experience can be improved because patients can get answered in time.

The most straightforward way to find similar questions is to retrieve questions that share the same keywords with the unsolved question. The drawback of such keyword-based approaches is that they only consider the information of the keywords and ignore the detailed information. In fact, in medical related questions, detailed information such as the patients' age, gender, and symptoms, is important and should not be missed when searching for similar questions. Therefore, the keyword-based approaches cannot provide satisfactory solutions in finding similar medical questions. Another popular approach is to calculate question-question similarity as the product of word-question similarities for the words in the other question \cite{zhang2016learning,jeon2005finding,xue2008retrieval,zhou2011phrase,berger2000bridging,cao2009use}. However, based on the collected dataset, we observe that people have very different ways to express the same concept in the medical/health domain, and thus the medical expressions can be quite diverse. For example, from the collected dataset, we found that there are in total $670$ ways to express the symptom ``headache" in Chinese medical questions. This exerts difficulties in estimating question similarities for the second approach.

In the light of these challenges, we propose a new solution for finding similar medical questions that can overcome the limitations of the existing methods. The essential part of the proposed method is to learn the fixed-length real-valued vector representations for medical questions by adopting Long Short-Term Memory (LSTM). The LSTM scans through each word in the question and automatically memorizes historical information. When the last word in the question is reached, the hidden state of LSTM is a vector representation for the input question, and it contains the semantic information of the whole question. Through LSTM, important details in the medical question can be encoded in the vector representations. Meanwhile, the vector representations are learned at the question level instead of word level, and thus similar questions with different expressions can share similar representations. Compared with existing approaches, the proposed method not only captures the important information contained in the medical question but also overcomes the challenge brought by the diversity of expressions.

To develop the end-to-end system for similar medical question finding, we further address two encountered practical issues. The first issue is the difficulty of obtaining large-scale training data. The core component of the system, LSTM, requires sufficient labeled training data. However, it will be time-consuming and expensive to manually create a large dataset containing similar medical question pairs. To solve this issue, we propose a solution to automatically generate similar medical question pairs based on random word replacing and dropping. The other issue is about efficiency. When a new unsolved question comes, we need to calculate the distance between this new question and all the existing questions to find similar ones. However, in practice, there are millions of answered questions in the corpus, and thus it would be time-consuming or even impossible to compute the distance between this new question and all the existing questions. To tackle this issue, we first extract some meta information such as question intention and question category from each question, and then select a small set of similar question candidates from an indexed question corpus. This strategy dramatically improves the efficiency and makes the developed system practical for real-world applications.

To evaluate the effectiveness of the developed system, we conduct experiments on a large-scale medical Q\&A dataset collected from \textit{baobaozhidao}. Experimental results show that the proposed method can find similar medical questions that are confirmed by human annotations. 
Furthermore, we demonstrate that the medical question vectors learned by the proposed method can capture important semantic information about the medical question, and can overcome the diversity issue of medical expressions.

To summarize, the contributions of this paper are as follows:
\begin{itemize}
\item  We propose an end-to-end system to solve a major challenge in crowdsourced medical Q\&A websites. Once a new question is posted, similar questions can be automatically found effectively and efficiently. This will give patients feedback quickly and expand the impact of online medical services.

\item The proposed system can overcome the limitations of the existing approaches when applied to medical Q\&A scenarios. In the proposed method, vector representations are learned at the question level, and the detailed information of medical questions are embedded. This enables accurate matching between the newly posted medical question and the solved questions. 

\item Experiments are conducted upon a large-scale real-world Q\&A dataset, and the results demonstrate the effectiveness of the proposed system.
\end{itemize}

In the following sections, after giving an overview of the proposed system, we present the details of each component in Section \ref{sec:method}. Then in Section \ref{sec:exp}, we demonstrate the effectiveness of the proposed system, which is tested on a large-scale real-world medical Q\&A dataset. We briefly discuss the related work in Section \ref{sec:related}, and finally conclude the paper in Section \ref{sec:conclusions}.

%% file: subfile/2_methodology.tex
\section{Overview of the System}

\begin{figure*}[tbh]
\centering
\includegraphics[width=0.9\textwidth]{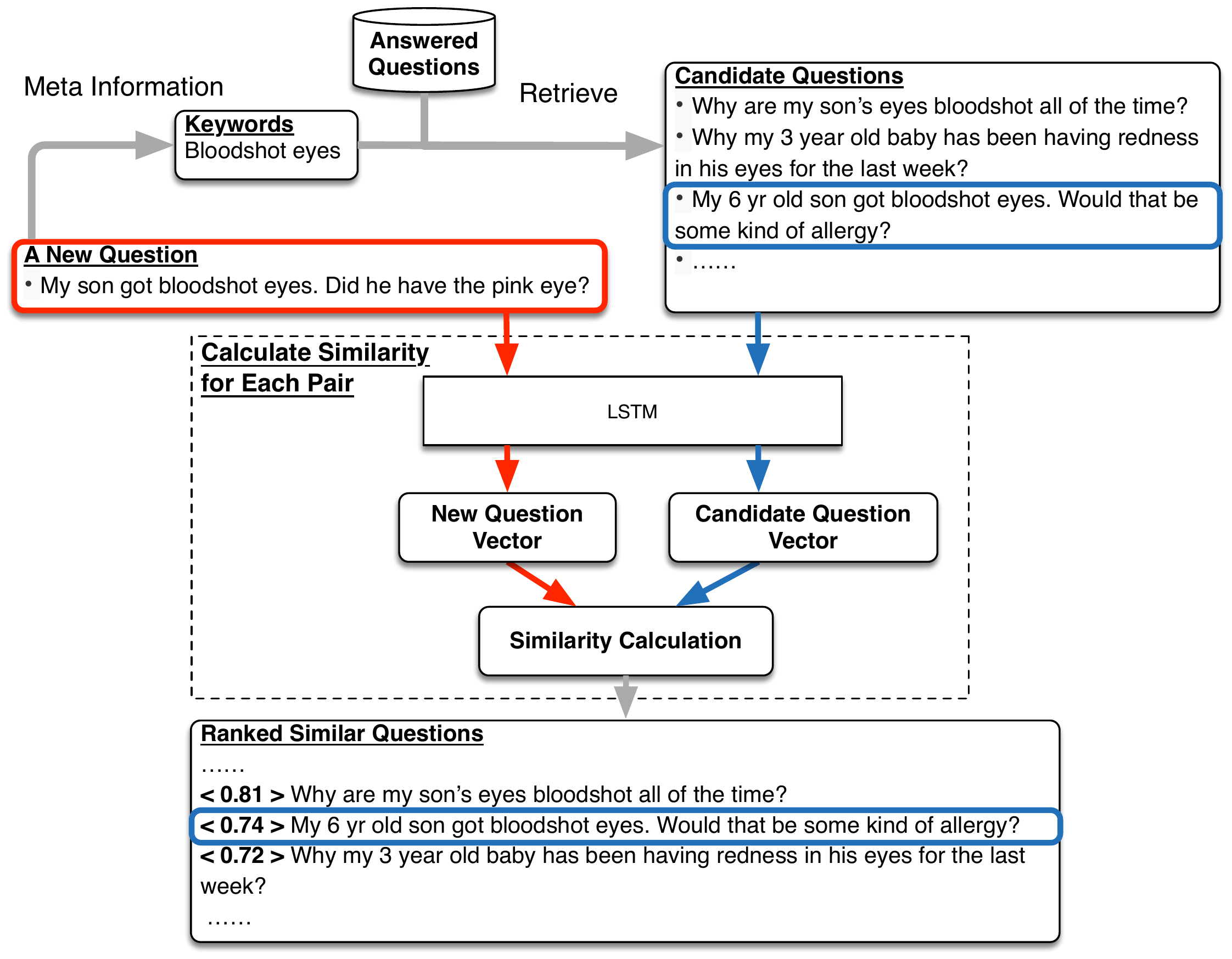}
\caption{System Overview.}
\label{overview}
\end{figure*}

To better illustrate the proposed system, we first give a high-level overview. Figure \ref{overview} shows the flow of the whole system that is designed to find similar questions for unsolved medical questions on crowdsourced medical Q\&A websites. When a new question, such as \textit{``My son got bloodshot eyes. Did he have the pink eye?"}, is posted on the website, we first extract some meta information such as question intention and question category from it. Such meta-information is helpful in identifying the most important word(s) in the unsolved question. Then we use the identified important word(s) to retrieve a small set of candidate questions from the existing question corpus. This step can significantly enhance the efficiency of the proposed system. Otherwise, we have to compare the new question with every solved question in the corpus, which is time-consuming or even impossible when the size of question corpus is large. Next, for each question in the candidate set, we calculate its similarity score with the unsolved question based on question vector representations produced by a trained LSTM. Finally, all the candidate questions are ranked according to their corresponding similarity scores, and the top ones whose similarity scores are larger than a threshold are considered as similar medical questions for the unsolved question. The details of each component are described in the following sections.

\section{Methodology}
\label{sec:method}

In this section, we first formally define the task. Then we introduce the core component, that is medical question vector representation learning, in the proposed system. Further, we discuss how to address two important issues, namely the training data generation issue and the efficiency issue. All these components make the proposed system practical in real-world applications.

\subsection{Task Definition}
We first introduce some notations that are used in this paper and then define the task. Let $\mathcal{X}_m = \left\lbrace \mathbf{x}_1, \mathbf{x}_2, ..., \mathbf{x}_{T_m} \right\rbrace$ denote the word embedding form of the $m$-th unsolved medical question, where $\mathbf{x}_t$ is the word embedding vector of the $t$-th word in this question, and $T_m$ is the total number of words in the $m$-th question. Our task is to automatically find the most similar questions from answered question corpus for the unsolved medical questions.

\subsection{Core Component}
\label{subsec:Proposed Model}
In this section, we describe the core component of the proposed system: how to learn the vector representations of medical questions, which are used for similarity calculation between a new question and the existing solved questions.  We first introduce LSTM which is adopted to learn vector representations of questions, and then we present the proposed architecture to find similar questions.

\subsubsection{LSTM}
\begin{figure}[tbh]
\centering
\includegraphics[width=0.48\textwidth]{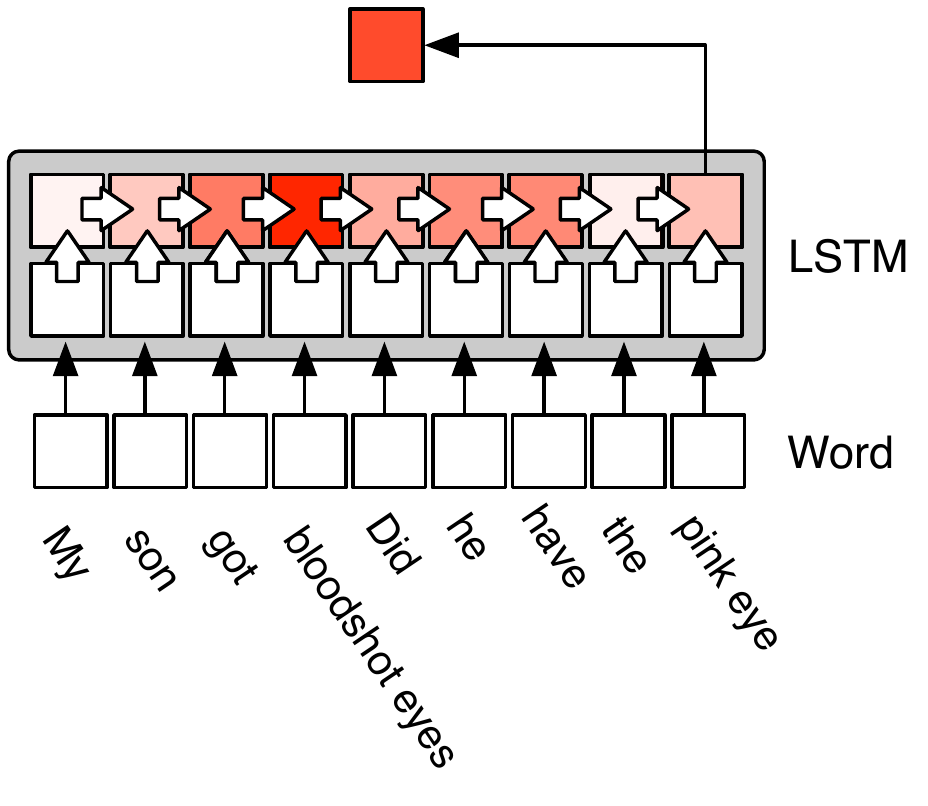}
\caption{LSTM can represent a question as a fixed-length real-valued vector.}
\label{lstm_vector}
\end{figure}

LSTM \cite{hochreiter1997long} is a popular recurrent neural network (RNN) architecture and it has been successfully applied to various applications \cite{sutskever2014sequence,bahdanau2014neural,graves2013speech}. In this paper, we adopt LSTM to learn the vector representations of the medical questions. When LSTM goes through the whole question, the information of each word in the sentence can be accumulated in its memory cell, and the hidden layer outputs the semantic representation of the question when LSTM reaches the end of the question. Thus the output vector representations of LSTM can capture the detailed information in the medical question at the sentence level. Figure \ref{lstm_vector} illustrates how LSTM represents a question as a fixed-length real-valued vector.

A typical LSTM cell contains an input gate, a forget gate, and an output gate. The input gate $\mathbf{i}$ controls how much information from the input can flow into the network; The forget gate $\mathbf{f}$ controls what historical information is unimportant and should be forgotten; The output gate $\mathbf{o}$ decides which part of the cell state $\mathbf{c}$ is allowed to output. The element values of the input gate, forget gate, and output gate range from $0$ to $1$. If an element value of a gate is $0$, then no information at the corresponding dimension can flush into the network; If the element value is $1$, then the entire information at the corresponding dimension can flow into the network. 

Formally speaking, the input of the LSTM is a vector sequence $\mathcal{X} = (\mathbf{x}_1, \mathbf{x}_2,\dots , \mathbf{x}_T)$, where $\mathbf{x}_t$ is the word embedding vector of the $t$-th word in the unsolved question. LSTM maps the input sequence $\mathcal{X}$ to the hidden vector $\mathbf{h}$ by sequentially applying the following equations on each word:

\begin{equation}
\label{LSTM cell}
\begin{split}
\mathbf{i_t} &= \sigma(\mathbf{W}_{ix}  \mathbf{x}_{t} + \mathbf{W}_{ih}\mathbf{h}_{(t-1)}+ \mathbf{b}_i);\\
\mathbf{f_t} &= \sigma(\mathbf{W}_{fx}  \mathbf{x}_{t} + \mathbf{W}_{fh}\mathbf{h}_{(t-1)}+ \mathbf{b}_f);\\
\mathbf{o_t} &= \sigma(\mathbf{W}_{ox}  \mathbf{x}_{t} + \mathbf{W}_{oh}\mathbf{h}_{(t-1)}+ \mathbf{b}_o);\\
\mathbf{c_t} &= \mathbf{c_{t-1}}\odot \mathbf{f_t}+ \mathit{tanh}\left(\mathbf{W}_{cx}\mathbf{x}_{t} + \mathbf{W}_{ch}\mathbf{h}_{t-1}+ \mathbf{b}_c\right)\odot \mathbf{i_t};\\
\mathbf{h_t} &=\mathit{tanh}\left( \mathbf{c_t} \right)\odot \mathbf{o_t},\\
\end{split}
\end{equation}

where 
\begin{itemize}
\item $\mathbf{i_t}$, $\mathbf{f_t}$, $\mathbf{o_t}$, $\mathbf{c_t}$, and $\mathbf{h_t}$ are the input gate, forget gate, output gate, cell state, and hidden state at the $t$-th word respectively. All of these vectors have the same dimension.
\item $\mathbf{W}$ and $\mathbf{b}$ are parameter matrix and bias vector.
\item $\sigma$ is the sigmoid function.
\item $\odot$ denotes the element-wise product of vectors.
\item $\textit{tanh}$ is the hyperbolic tangent function. 
\end{itemize}

The final hidden state vector $\mathbf{h_{T}}$ can be regarded as the semantic representation of the whole question. For clarity, we use $\mathbf{v}$ to denote the question vector representation, i.e. $\mathbf{v} = \mathbf{h_T}$.

\subsubsection{Similarity Calculation} 
\label{sub:architecture}
Before describing how to calculate the similarity score for a pair of questions, we further introduce some notations. Let $\mathbb{C}_m = \left\lbrace \mathcal{C}_{m,1}, \mathcal{C}_{m,2}, ... \right\rbrace$ denotes the candidate question set for the $m$-th unsolved question, where the element $\mathcal{C}_{m,n}$ is the word embedding form of the $n$-th candidate question. Our goal is to automatically find similar questions from answered question corpus, i.e., rank the candidate questions in $\mathbb{C}_m$ according to their similarity scores with the unsolved question.

\begin{figure}[tbh]
\centering
\includegraphics[width=0.5\textwidth]{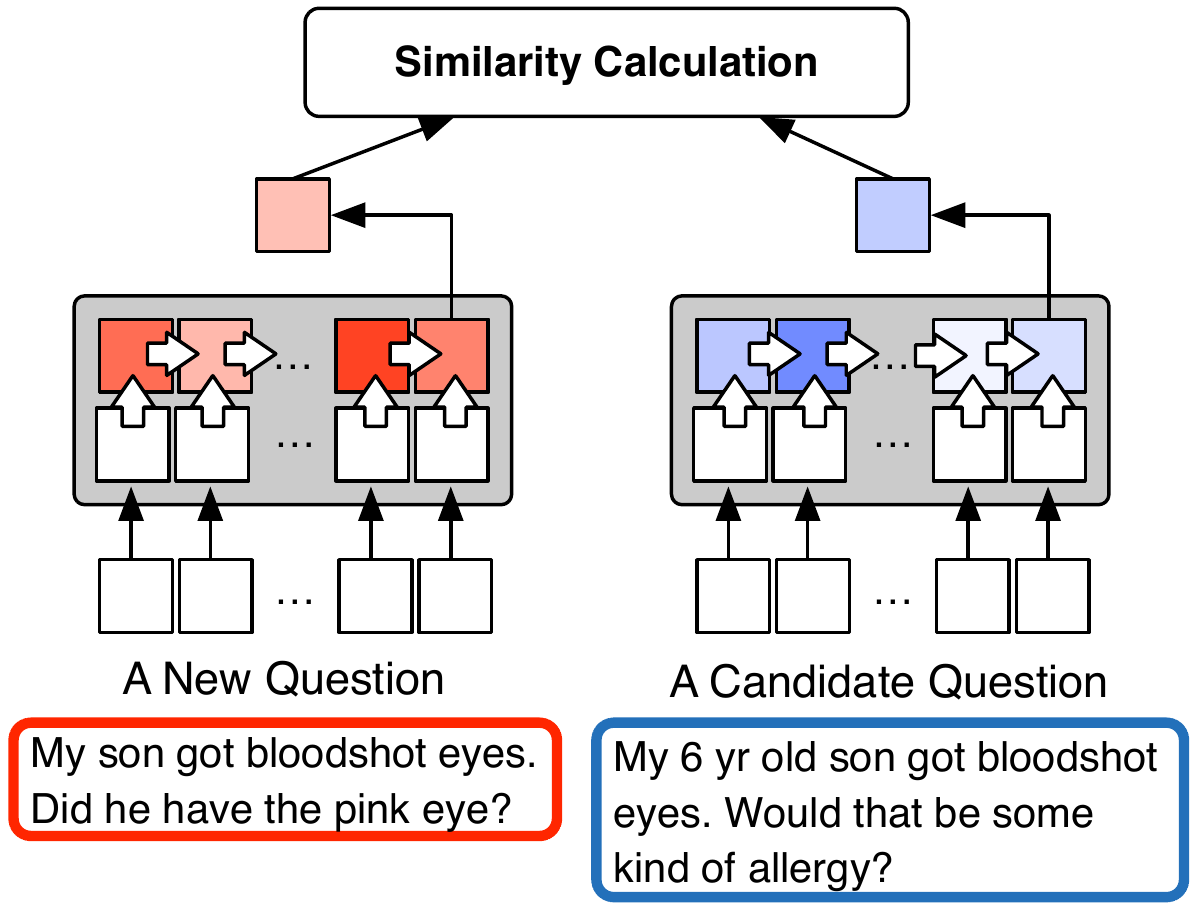}
\caption{The architecture for similarity calculation.}
\label{architecture}
\end{figure}

Given the $m$-th unsolved question $\mathcal{X}_m$ and one of its candidate question $\mathcal{C}_{m,n}$, a pre-trained LSTM (we will discuss the pre-training of LSTM in the following section) is used to map both the unsolved question and the candidate question into vectors $\mathbf{v}_{X_m}$ and $\mathbf{v}_{C_{m,n}}$, where $\mathbf{v}_{X_m}$ and $\mathbf{v}_{C_{m,n}}$ are the final hidden states of LSTM when the input sequence is $\mathcal{X}_m$ and $\mathcal{C}_{m,n}$ respectively. Then the similarity score between the unsolved question $\mathcal{X}_m$ and the candidate question $\mathcal{C}_{m,n}$ is defined as $S(\mathcal{X}_m, \mathcal{C}_{m,n}) = <\mathbf{v}_{X_m},\mathbf{v}_{C_{m,n}}>$,
where $<\cdot , \cdot>$ is the inner product of two vectors. Figure \ref{architecture} shows the proposed architecture.

Then all the candidate questions can be ranked according to their similarity scores with the unsolved question. The higher the similarity score is, the more similar the candidate question is to the unsolved question.

\begin{table*}[tbh]
\centering
\begin{tabular}{cc}
\toprule
&\begin{CJK*}{UTF8}{gbsn}四个月的宝宝发烧拉肚子怎么办?\end{CJK*} (Four-month-old baby is suffering from fever and diarrhea. What to do?) \\
Negative & \begin{CJK*}{UTF8}{gbsn}怀孕能吃辣的东西吗?\end{CJK*} (Can I eat spicy food during pregnancy?)\\
\cmidrule{2-2}
Samples &\begin{CJK*}{UTF8}{gbsn}宝宝鼻塞好不了怎么办?\end{CJK*} (My baby cannot get rid of stuffy nose. What can I do?) \\
& \begin{CJK*}{UTF8}{gbsn}胎心监护显示波浪明显变异是什么意思?\end{CJK*} (The wavelet displayed by fetal heart monitor has obvious variation. What it means?) \\
\midrule
&\begin{CJK*}{UTF8}{gbsn}宝宝免疫低该咋办，着急\end{CJK*} (I am so worried that my kid has low immunity. What should I do?) \\
Positive & \begin{CJK*}{UTF8}{gbsn}孩子免疫低下咋办，着急\end{CJK*} (I am so worried that my kid has poor immunity. What should I do?)\\
\cmidrule{2-2}
Samples &\begin{CJK*}{UTF8}{gbsn}宝妈睡觉左侧好还是右侧好?\end{CJK*} (Which side to sleep is better for expecting mother, right or left?) \\
& \begin{CJK*}{UTF8}{gbsn}孕妇睡觉左侧好右侧好?\end{CJK*} (Which side to sleep is better for pregnant women
, right, left?)\\
\bottomrule
\end{tabular}
\vspace{0.1in}
\caption{Generated training data samples.}
\label{generated_data_sample}
\end{table*}

\subsubsection{Pre-training of LSTM} 
As mentioned above, the final hidden state of LSTM can be regarded as a summarized representation of the input sentence. However, to make such representations meaningful, the training procedure of LSTM should be guided by properly labeled data.  

Let $y\in \left\lbrace 0, 1 \right\rbrace $ denote the binary label for a question pair. If two questions are similar to each other, then the label for this question pair should be $1$; otherwise, it is $0$. Each record in the training data contains one question pair and its corresponding label, and the $j$-th record is denoted as $\left\lbrace \mathcal{Q}_j,\mathcal{Q}_j^{'}\right\rbrace$ and $y_j$. Then the loss function to train the LSTM network is defined as follows:
\begin{equation}
l = \sum_{j = 1}^{J}\left[S\left(\mathcal{Q}_j,\mathcal{Q}_j^{'}\right) - y_{j}\right]^2,
\label{lstm_loss}
\end{equation}
where $J$ is the total number of question pairs in the training dataset. The loss function is the sum of the Euclidean distances between the estimated similarity scores and their corresponding true labels. By minimizing the loss function, the network is trained to make the similarity score of each question pair as close to their true label as possible. If the input question pair is a positive sample, then the estimated similarity score should be close to $1$; otherwise, the estimated similarity score should be close to $0$. Guided by the training data, LSTM can learn how to project medical questions into real-valued vectors such that the questions that share similar semantic meanings will have similar representations.

\subsection{Training Data Generation}
\label{sub: Training Data Generation}

Due to the nature of deep neural networks, sufficient training data is required to train an LSTM. However, it would be costly if we manually create a large-scale training data about medical questions. In order to solve this issue, we propose an alternative way to generate both positive and negative labeled training data automatically.

Similar to negative sampling \cite{mikolov2013distributed}, negative training data is generated by randomly sampling pairs of questions from the question corpus. We further impose a constraint that the sampled negative pairs should not share any common medical entities. With this strategy, the sampled negative question pair can be labeled as $0$, and the label is correct with a high probability.

For positive question pairs, it seems unrealistic to directly sample them from question corpus as finding similar questions from the question corpus is in fact the goal of the proposed system.
To tackle this problem, we propose a solution to generate similar question pairs by replacing and dropping words randomly. For each word in a medical question, it will be replaced by one of its synonyms with a predefined probability based on a medical dictionary. In general, randomly replacing some words by their synonyms does not change the meaning of the whole sentence. 
However, semantically similar questions can be more than sharing similar words. It is also likely that two questions have similar semantic meanings though they have different expressions. In order to mimic this, random word dropping is also applied. We set a probability, such as $0.1$, to randomly drop each word in a medical question. For example, given one question, \textit{``My baby has a cold with nose keeping flowing. What can I do?"}, the newly generated similar question after the two steps can be \textit{``Kid has a cold with a runny nose. What can do?''}.

\begin{table*}[h]
\centering
\begin{tabular}{cc}
\toprule
Medical Question & Identified Keywords \\
\midrule
\begin{CJK*}{UTF8}{gbsn}14个月小孩子有O形腿，怎么矫正？\end{CJK*} &\begin{CJK*}{UTF8}{gbsn}O型腿 矫正\end{CJK*} \\
14-month-old baby has bow legs. How to correct it? &  bow legs correction \\
\begin{CJK*}{UTF8}{gbsn}52天女婴频繁吐奶，怎么办？\end{CJK*} &\begin{CJK*}{UTF8}{gbsn}吐奶\end{CJK*} \\
52-day-old baby girl frequently vomits milk. What should I do? & vomit milk \\
\begin{CJK*}{UTF8}{gbsn}小孩远视，散光，该怎么办？\end{CJK*} &\begin{CJK*}{UTF8}{gbsn}远视散光\end{CJK*} \\
The kid has farsightedness and astigmatism. What should I do? & farsightedness  astigmatism\\
\bottomrule
\end{tabular}
\vspace{0.1in}
\caption{Examples about keyword identification based on the question meta information.}
\label{meta_information_example}
\end{table*}

Some generated negative and positive question pairs are shown in Table \ref{generated_data_sample}. Since the original questions are in Chinese, we also provide corresponding English translations. From these sampled examples, we observe that the generated question pairs are meaningful. If LSTM is trained with only the generated positive samples, it may be over-fitted and only learn knowledge about synonyms; However, by fitting both generated positive and negative pairs, LSTM can automatically learn how to represent medical sentences as meaningful vectors, which is more than the knowledge about synonyms. We will experimentally confirm this in Section \ref{subsec:case} and \ref{subsec:vector}, and demonstrate the usefulness of the generated training data.

\subsection{Improving Efficiency}
\label{subsec: Improving Efficiency}
When the size of question corpus is large, computing the distance between the new question and every solved question in the corpus will be time-consuming or even impossible. In our experiments, the collected dataset contains around $10$ million solved questions. Thus it is infeasible to compare a new question with every solved question. Therefore, to improve the efficiency, we propose to select a small set of candidate questions first, and then only compute the distance between the new question with the candidate questions in the set.

In order to preselect a candidate question set, using some keywords in the new question can be a good strategy. As mentioned earlier, even if the keyword based approaches cannot find similar questions precisely, they are still helpful in filtering out totally unrelated questions. Then the next difficulty is: Which word(s) in the new question should be selected to filter the whole question corpus? Let's take the question \textit{``My son got bloodshot eyes. Did he have the pink eye?''} as an example. Considering it is a medical question, the medical terms \emph{bloodshot eyes} and \emph{pink eye} can be potential keywords. However, a medical question usually contains many medical terms, and if we select candidate questions that contain all these medical terms, the candidate set is probably to be empty. Meanwhile, not all the medical terms in the new questions are equally important. Consider the above question: this user's son has the symptom of \emph{bloodshot eyes}, and she/he is asking questions about possible diseases. \emph{Pink eye} is this user's guess, so it should not be selected as a keyword. Inspired by these observations, we propose to choose keywords in the new medical questions by its meta information such as question intention and question category. Let's revisit the example: this new question is about children-health, and it intends to determine possible diseases accordingly to the given symptoms. Based on the meta information, we can retrieve candidate questions that contain keyword \emph{bloodshot eyes} and have the same question category (children-health in this example) and question intention (seeking diagnosis based on symptoms in this example). Usually, this will only give tens to hundreds of candidate questions, which can dramatically improve efficiency. 

For most crowdsourced medical Q\&A websites, question category information is available. For question intention detection, we adopt the method described in \cite{zhang2016mining}. In Table \ref{meta_information_example}, we give some medical questions and the corresponding keywords that are identified based on the meta information of questions. From these examples, we can observe that question meta information can help us to accurately identify keywords, and these keywords will be used to filter the whole question corpus to create a small candidate question set.

%% file: subfile/3_exp.tex
\section{Experiments}
\label{sec:exp}
In this section, we evaluate the proposed system with a large-scale medical Q\&A dataset, and demonstrate its effectiveness from the following aspects: 
(1) Given a new unsolved question, the proposed system can find similar questions efficiently and precisely.
(2) Through some case studies, we observe that the proposed system can find similar questions even when they have different expressions.
(3) The learned vector representations for questions can automatically encode the importance of each word in the medical question.

\begin{table*}[tbh]
\centering
\begin{tabular}{cc}
\toprule
{\color{blue}New Question} & {\color{blue}\begin{CJK*}{UTF8}{gbsn}小孩缺锌吃什么食物和补锌的药? \end{CJK*} Kid zinc deficiency, what food and zinc supplements to eat? } \\
&\begin{CJK*}{UTF8}{gbsn}小孩缺锌吃什么补锌?\end{CJK*} Kid zinc deficiency, what to eat to supply zinc?  \\
Similar Questions &\begin{CJK*}{UTF8}{gbsn}儿童缺锌吃什么食品如何补锌?\end{CJK*} Zinc deficiency in child. What kind of food to eat? How to supply zinc? \\
 &\begin{CJK*}{UTF8}{gbsn}缺锌吃什么最补锌?\end{CJK*} Zinc deficiency, what is the best to eat to supply zinc?\\
\midrule
{\color{blue}New Question} & {\color{blue}\begin{CJK*}{UTF8}{gbsn}宝宝鹅口疮反复发作，怎样根治?\end{CJK*}
\small{The child suffers from recurrent bouts of thrush. How to cure it completely?}}\\
\multirow{2}{*}{Similar Questions}&\begin{CJK*}{UTF8}{gbsn}宝宝鹅口疮好了怎么又复发？该怎么办?\end{CJK*} Why the thrush of my baby relapses. What should I do? \\ 
 &\begin{CJK*}{UTF8}{gbsn}小孩鹅口疮怎么治疗，好了又复发！\end{CJK*} How to treat thrush in kid? The thrush relapses.\\
\midrule
{\color{blue}New Question} &{\color{blue}\begin{CJK*}{UTF8}{gbsn}烫染头发后还可以喂奶吗？\end{CJK*} After I curl and dye hair, can I  breastfeed my baby?} \\
 &\begin{CJK*}{UTF8}{gbsn}喂奶期间可不可以烫染头发？\end{CJK*} During the breastfeeding period, can I curl and dye hair or not? \\
Similar Question &\begin{CJK*}{UTF8}{gbsn}哺乳期间可以烫染头发吗？\end{CJK*} During the lactation period, can I curl and dye hair?  \\
 &\begin{CJK*}{UTF8}{gbsn}喂奶期间能烫或染头发吗？\end{CJK*} During the  breastfeeding period, can I curl or dye hair?\\
\bottomrule
\end{tabular}
\vspace{0.1in}
\caption{Case Studies.}
\label{example:similar_questions}
\end{table*}

\subsection{Experiment Setup}
\subsubsection{Data Collection}
The question corpus is collected from \textit{baobaozhidao}\footnote{\url{baobao.baidu.com}}, one of the biggest crowdsourced Q\&A website about maternal and child health in China. On this Q\&A forum, mothers can post medical or health-related questions about their children or themselves, and other users, either registered doctors or experienced mothers, will contribute their answers. However, most of the new questions cannot get answers quickly. Thus it will be helpful if the proposed system finds some similar questions from solved question-answer pairs. To fulfill this goal, we crawled totally $10,450,825$ questions from this medical Q\&A website.

\subsubsection{Training of LSTM}
As aforementioned, training of LSTM needs sufficient labeled data. To generate large-scale labeled training data, we first randomly select $865,945$ questions from the question corpus. Then for each question, one positive pair and three negative pairs are generated following the strategies described in Section \ref{sub: Training Data Generation}. Eventually, we get a training dataset containing $3,463,780$ labeled question pairs.

These generated raw question pairs cannot be directly fed into LSTM, and we need to transform the question texts into sequences of word vectors. In order to learn word vectors, we use all the collected question texts. Skip-gram architecture in Word2vec package\footnote{\url{https://code.google.com/archive/p/word2vec/}} is adopted to train word vector representations. The dimensionality of the learned vectors is set to be $100$, the context window size is set to be $8$, and the minimum occurrence count is set to be $5$. For more details, please refer to \cite{mikolov2013distributed}. By feeding the word embedding forms of question pairs, LSTM is trained to minimize the loss defined in Eq. (\ref{lstm_loss}). The pre-trained LSTM will be used to transform medical question texts into question vectors.

\subsection{Case Studies}
\label{subsec:case}
In order to have an intuitive impression of the proposed system, we randomly select some new questions and their similar questions identified by the proposed system. Table \ref{example:similar_questions} shows some examples. Since both the unsolved questions and the identified similar questions are in Chinese, we also provide their English translations.

From the perspective of semantic expressions, the similar questions identified by the proposed system for these three cases have different patterns.

In the first case, the identified similar questions have similar expression structure with the unsolved question. Compared with the unsolved question, the identified similar questions substitute synonyms at the word level or drop some words: replacing \textit{``kid"} with \textit{``child"}, or dropping word \textit{``kid''}. Finding such kind of similar questions is relatively easy.

The second case is slightly different from the first case. Similar to the first case, we also observe some synonym substitutions, such as replacing \textit{``child''} with \textit{``kid''} or \textit{``baby''}. But more importantly, we observe that even some phrases are very different, the proposed system still can recognize that they express the same meaning, such as \textit{`` \begin{CJK*}{UTF8}{gbsn}反复发作\end{CJK*} ( recurrent bouts)"}  and \textit{`` \begin{CJK*}{UTF8}{gbsn}好了又复发\end{CJK*} (relapse)"}. Meanwhile, the expression structures of the identified similar questions are different. In the new question,\textit{ ``recurrent"} comes first, followed by \textit{``how to cure"}, while in the identified similar questions, the order can be reversed. This indicates that our system can not only find questions sharing synonymous words but also find semantically similar questions regardless of the difference in the expressions. 

The last case is more interesting. The identified similar questions and the unsolved question have very different expressions. The unsolved question mentions whether she can nurse her baby after she curls and dyes hair, and the identified similar questions ask whether she can curl and dye hair during the lactation period. This implies that the proposed system can find semantically similar questions even when these questions have different expressions.

The above three cases demonstrate that the proposed system is able to find similar questions more than just at the lexical level. Similar questions that have different sentence structures, or different expressions with the unsolved question can also be identified by the proposed system.

Moreover, we also observe that for some unsolved questions, there are no identified similar questions. For example, \textit{`` \begin{CJK*}{UTF8}{gbsn}请问小孩晚上醒来总说看到很多红色、绿色点点是怎么回事？\end{CJK*}(The kid always complained of seeing many red and green dots after wakeup at night. What is wrong?)"} and \textit{``\begin{CJK*}{UTF8}{gbsn}5岁女孩长期头痛和左后臀痛和腹痛，怎么办？\end{CJK*}(The five-year-old girl has long-term pain in head, left butt, and abdomen. What to do?)"}. The reason is that these cases are so unique that all the questions in the candidate set are different from them. The results follow our intuition: the symptom in the first question is too rare and the second question is too specific. Therefore, not finding similar questions for some cases is also desired: the proposed system is able to distinguish unique questions and output no results rather than misleadingly identify some ``similar'' questions.

\subsection{Quantitative Evaluation}
In the case studies section, we provide some case studies to demonstrate the effectiveness of the proposed system. Here we further conduct quantitative evaluations to confirm it. We first describe the collection of ground truth, and then introduce two metrics \textbf{Precision} and \textbf{Rank Correlation} that are used to evaluate the performance of the proposed system. Experimental results under these performance metrics are reported and further analyzed.

\subsubsection{Ground truth Collection}
\begin{figure}[tbh]
\centering
\includegraphics[width=0.49\textwidth]{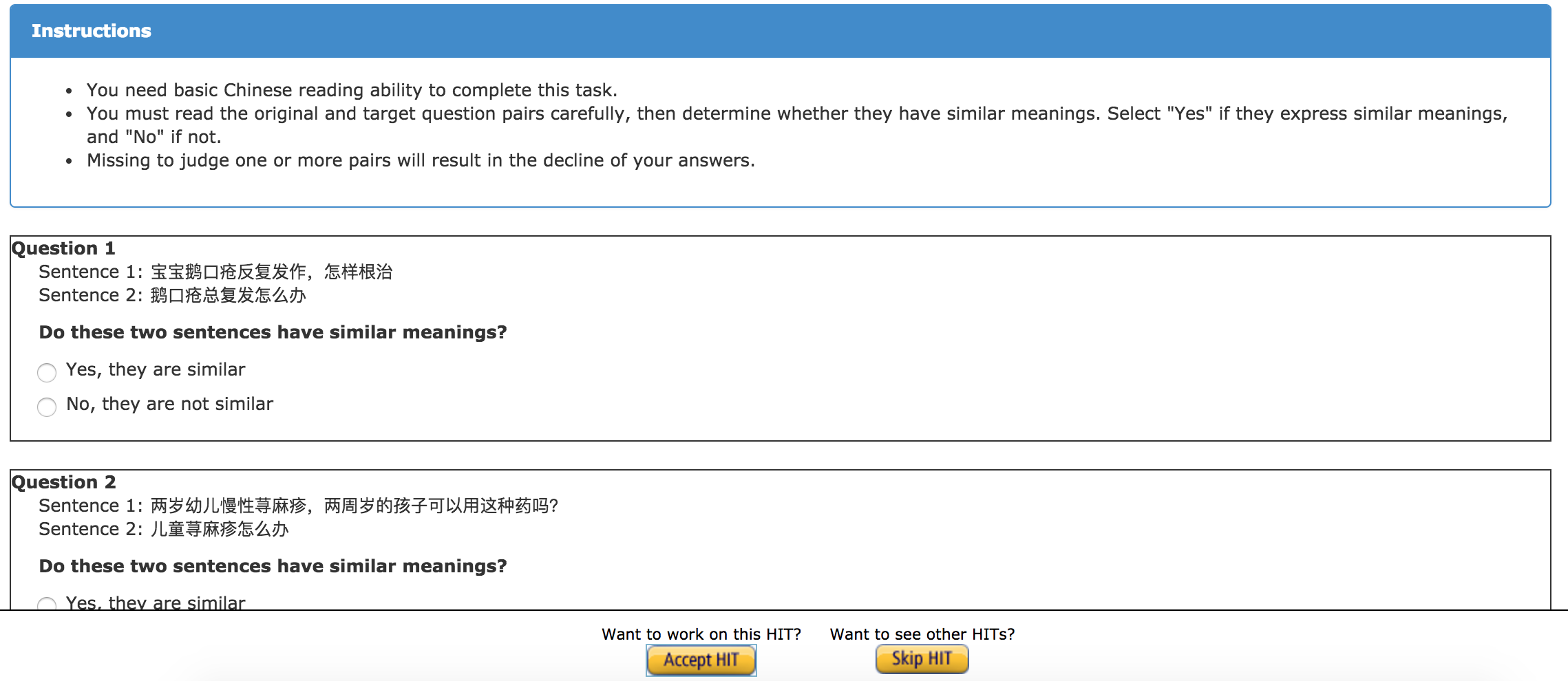}
\caption{A screenshot of the posted annotation task on Amazon Mechanical Turk.}
\label{aws}
\end{figure}

For the purpose of quantitative evaluation, ground truth labels from human annotators are required. Due to the labeling cost, we randomly select $100$ new questions from \textit{baobaozhidao}, and the proposed system identifies corresponding top-$5$ similar questions. We post the identified similar questions on Amazon Mechanical Turk (AMT)\footnote{\url{https://www.mturk.com/mturk/welcome}} for ground truth collection. Each micro-task contains $10$ pair of questions, where each pair contains a new medical question and a similar question identified by the proposed system. 
For example, `` \textit{\begin{CJK*}{UTF8}{gbsn}小孩鹅口疮怎么治疗，好了又复发\end{CJK*} (How to treat thrush in my kid? The thrush relapses)} " and ``\textit{\begin{CJK*}{UTF8}{gbsn}宝宝鹅口疮反复发作，怎样根治\end{CJK*} (The child suffers from recurrent bouts of thrush. How to cure it completely?)} " is a pair of questions. 
Workers on AMT are asked to evaluate whether these pairs of medical questions are similar or not. Each micro-task is labeled by $20$ workers. Figure \ref{aws} gives a screenshot of our posted tasks on AMT. 

To ensure the quality of human annotation, we impose two question pairs that we know the ground truth labels into each micro-task. If a worker gives different answers to these two question pairs, then we will not accept his answers in this micro-task.
Finally, $8,770$ labels are collected for ground truth construction.

\subsubsection{Result Analysis}
Based on these collected labels from AMT workers, we want to answer the following two questions: (1) Are the identified similar medical questions truly similar to the unsolved questions? (2) How similar the identified questions are to the unsolved questions? For the purpose of quantitative evaluation, we adopt \emph{precision} and \emph{rank correlation} to answer the above two questions.

\textbf{Precision.} Performance metric \emph{precision} is adopted in order to verify whether the identified questions are similar to their corresponding unsolved questions or not. For each pair of medical questions, i.e., an unsolved question and a similar question identified by the proposed system, ground truth label can be assigned based on the portion of workers labeling it as \textit{``similar"}. For one question pair, if the majority of the workers (i.e., more than $50\%$ of the workers) consider this pair as similar, then the ground truth is \textit{``similar"}; otherwise, its corresponding ground truth label will be \textit{``not similar"}.
Then the precision can be defined as the portion of question pairs whose ground truth is \textit{``similar''} among all the question pairs. The bigger the precision is, the more identified questions are similar to their corresponding unsolved questions. The precision of the proposed system is $87.98\%$, which means that $87.98\%$ of the identified questions are considered as ``similar'' by the human workers.

\begin{figure}[tbh]
\centering
\includegraphics[width=0.49\textwidth]{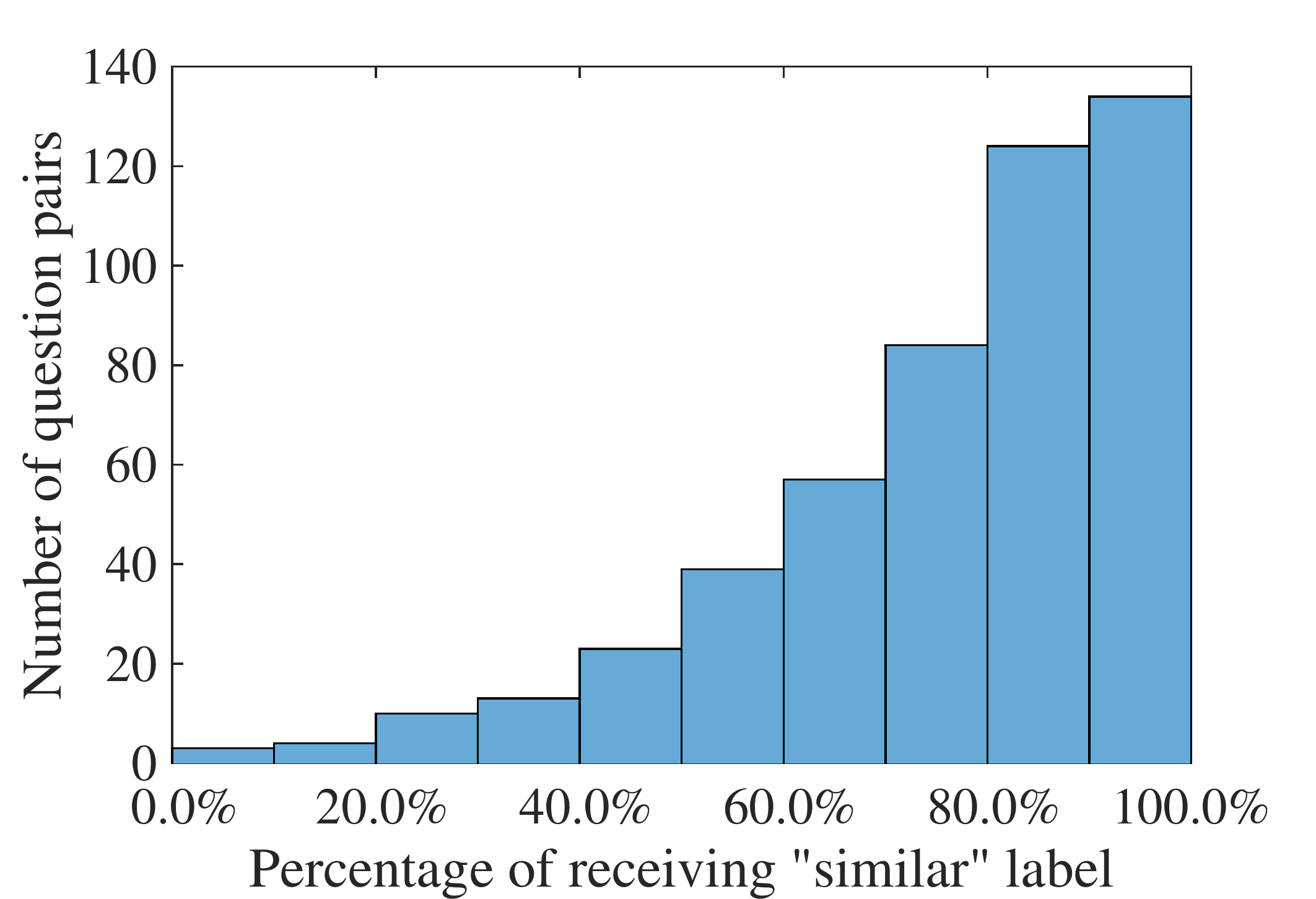}
\caption{The histogram about the percentage of receiving \textit{``similar"} labels for question pairs.}
\label{histogram}
\end{figure}

From the labels provided by workers, we also observe that even when two different questions pairs are assigned the same ground truth label, \textit{``similar"}, the percentage of workers labeling them as \textit{``similar"} may be different. Suppose for one question pair, there are only $11$ out of $20$ workers labeling it as \textit{``similar''}, while for another question pair, there are $19$ out of $20$ workers labeling it as \textit{``similar"}. The confidence degrees of the similarity about these two question pairs are obviously different, but the precision ignores this information. Thus, we further plot a histogram to examine the ``strength'' of the similarity. Figure \ref{histogram} shows that most of the question pairs returned by the proposed system are agreed to be similar by more than $80\%$ of the workers, which indicates that the proposed system indeed can provide similar questions that are meaningful and reasonable. Figure \ref{histogram} also motivates us to adopt the following performance metric, rank correlation.

\textbf{Rank Correlation.} The proposed system outputs the similarity scores that contain similarity degree information for each identified similar question. Though human workers cannot easily output a similarity score for a question pair, the percentage of positive labels implies the similarity degrees as shown in Figure \ref{histogram}. Therefore, we use this information to further quantitatively evaluate the proposed system by adopting rank correlation coefficients. For each unsolved question, we rank all the identified similar questions according to the corresponding portions of workers giving label \textit{``similar"}, which can be regarded as ground truth ranking order. Meanwhile, the proposed system also outputs these similar questions in a ranked list. So for each case, i.e., an unsolved medical question, there are two associated ranked lists that can be used to evaluate the identified similar question order.

\begin{table}
\centering
\small
\begin{tabular}{c c c }
\toprule
 Case & Kendall Coefficient    & Spearman's Coefficient\\
\hline
1    &    0.1515    &    0.1469\\
2    &    0.8000    &    0.9000\\
3    &    0.8000    &    0.9000\\
4    &    0.4000    &    0.4000\\
5    &    0.1818    &    0.2797\\
6    &    0.3333    &    0.1429\\
7    &    0.0769    &    0.0725\\
8    &    0.6667   &    0.8000\\
9    &    0.2191   &    0.3286\\
10    &    0.2500    &    0.1961\\
11    &    0.1111   &    0.0299\\
12    &    0.2000    &    0.3000\\
13    &    0.8476    &    0.9250\\
14    &    0.5000    &    0.6429\\
15    &    0.2281    &    0.2421\\
16    &    0.5333   &    0.7059\\
$\cdots$ & $\cdots$ & $\cdots$ \\
\bottomrule
\end{tabular}
\vspace{0.1in}
\caption{Rank Correlation Coefficients for Case Studies.}
\label{table: Rank Correlation Coefficient Statistics}
\vspace{-0.1in}
\end{table}

\begin{figure}[tbh]
\centering
\includegraphics[width=0.48\textwidth]{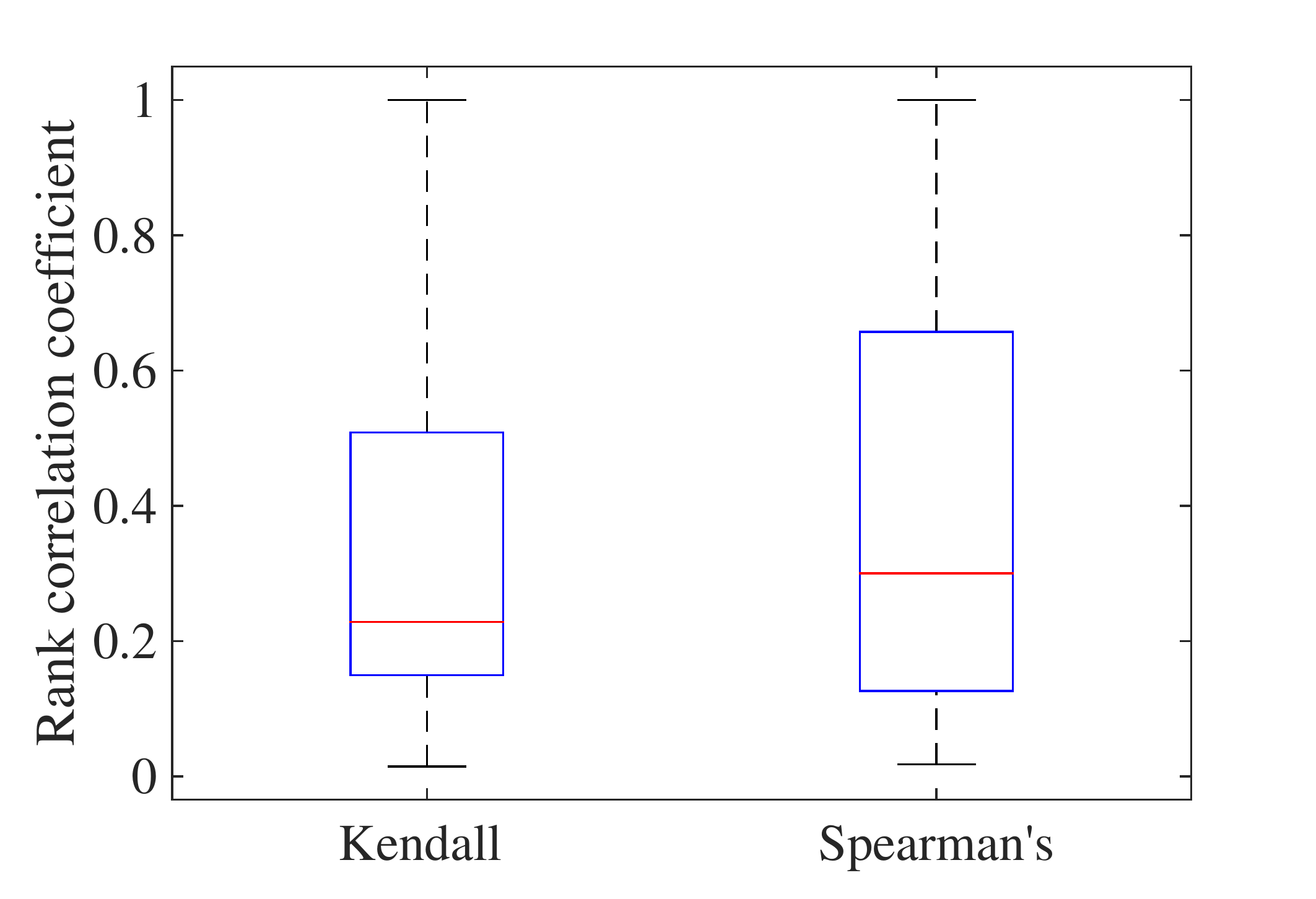}
\caption{Box plot of all rank correlation coefficients.}
\label{Boxplot}
\end{figure}

In order to measure the correlation between the ranked list given by the proposed system and the ground truth ranked list, we adopt two popular rank correlation coefficients, Kendall rank correlation coefficient\footnote{\url{https://en.wikipedia.org/wiki/Kendall_rank_correlation_coefficient}} and Spearman's rank correlation coefficient\footnote{\url{https://en.wikipedia.org/wiki/Spearman\%27s_rank_correlation_coefficient}}. The value of both two coefficients ranges from $-1$ to $1$. Positive coefficient values indicate that the compared two ranked lists are correlated positively, and negative values indicate negative correlations. The absolute values of the coefficients indicate the strength of correlation. The bigger value, the stronger correlation. For each case, we calculate the two rank correlation coefficients between the ranked lists given by the proposed system and the ground truth ranked list, and due to the space limit, only the first $16$ cases are reported in Table \ref{table: Rank Correlation Coefficient Statistics}. The distributions of calculated rank coefficients for all the cases are plotted in Figure \ref{Boxplot}.

From Table \ref{table: Rank Correlation Coefficient Statistics} and Figure \ref{Boxplot}, we observe that for most of the cases, both Kendall and Spearman's rank correlation coefficient are positive, which means that for the unsolved medical questions, the outputted ranked lists of similar questions given by the proposed system are positively confirmed by human labels. These results imply that the similarity scores calculated by the proposed system can accurately reflect how similar the identified question is to the corresponding unsolved medical question.

\subsection{Examination of Question Vector Representations}
\label{subsec:vector}
The core component of the proposed system is to learn vector representations for medical questions. Therefore, in this section, we examine the effectiveness of the learned vector representations for medical questions.

A good medical question vector representation should encode necessary information and be able to automatically attenuate the unimportant words and detect the salient words in the medical question. To verify that, we conduct the following experiment: We change some words in one medical question; Then transform the original question as well as the question with changed words into vector representations by applying the pretrained LSTM; Finally calculate the similarity between the questions with changed words and the original question. If the vector of the modified question is not similar to the vector of the original question, it indicates that the changed word is important to the whole question. Table \ref{sentence_vector} shows a case study with the original question ``\begin{CJK*}{UTF8}{gbsn}宝宝鹅口疮反复发作，怎样根治\end{CJK*}" (\textit{My child has recurrent thrush. How to cure it?}). We construct two changed questions: replacing the word \textit{``Child"} with \textit{``baby"}, and replacing the word \textit{``thrush"} with \textit{``cold"} respectively. Table \ref{sentence_vector} shows the vector similarities between the modified medical questions and the original medical question.

\begin{table}[tbh]
\centering
\begin{tabular}{cc}
\toprule
Modified Medical Questions & Vector Similarity \\
\midrule
\begin{CJK*}{UTF8}{gbsn}孩子鹅口疮反复发作，怎样根治\end{CJK*} & \multirow{2}{*}{0.99} \\
My baby has recurrent thrush. How to cure it? & \\
\midrule
\begin{CJK*}{UTF8}{gbsn}宝宝感冒反复发作，怎样根治\end{CJK*} & \multirow{2}{*}{0.52}\\
My child has recurrent cold. How to cure it? & \\
\bottomrule
\end{tabular}
\vspace{0.1in}
\caption{The question vector similarity between the modified medical questions and the original medical question \emph{``My child has recurrent thrush. How to cure it?"}}
\label{sentence_vector}
\end{table}

From the table, we observe that the vector similarity of the changed question with word \textit{``baby"} is quite close to $1$, which indicates that after synonymy substitution, the new question is similar to the original question. While the vector similarity of changed question with word \textit{``cold"} is far from $1$ as the word \textit{``cold"} dramatically changes the meaning of the whole medical question. This case study verifies that the proposed system can automatically detect salient words in the medical questions, and the medical question vector representations are meaningful.

%% file: subfile/4_related.tex
\section{Related Work}
\label{sec:related}

The problem of finding similar questions from Q\&A websites has attracted increasing attention since people realize that many posted questions cannot be answered timely\cite{li2010routing}. 
At the early stage, the methods people used to retrieve similar questions are inspired by the general information retrieve models \cite{ponte1998language,zhai2001study}. However, such retrieval-based methods only consider the information about keywords, but the detailed information in the questions is not captured. 
Then topic modeling based methods are developed to include more information extracted from questions \cite{ji2012question,cai2011learning,zhang2014question}. 
Later, the problem of finding similar questions is casted as a translation task, as a similar question can be regarded as a ``translation" version of the target question \cite{jeon2005finding,xue2008retrieval,zhou2011phrase,berger2000bridging,cao2009use}. Further, question category information \cite{cao2010generalized} and question topic information \cite{blei2003latent} are incorporated to enrich the translation-based methods.
More recently, people have begun to apply deep learning techniques to solve the task of finding similar questions \cite{das2016mirror,qiu2015convolutional,zhou2016learning,das2016together}. However, all these existing methods are not designed to deal with medical questions, in which some unique challenges are critical and need to be considered: The detailed information heavily affects the semantic meaning of medical questions; Patients tend to use diverse terms to express the same meaning; And a large-scale labeled training data is difficult to obtain. In this paper, we propose an end-to-end system to tackle these unique challenges, and the system can successfully find similar medical questions from Q\&A websites.

In terms of techniques, there are some existing methods to encode a sentence into a vector representation, such as \emph{para2vec} \cite{le2014distributed} and \emph{skip vector} \cite{kiros2015skip}. However, these methods rely on the context sentences of a target sentence to learn its sentence vector representation. While in the medical Q\&A scenario, most of the questions are short, and no context sentences are available. 
Recently, LSTM model has been explored to encode sentence-level information for some specific domains \cite{palangi2016deep,wan2015deep}. In the proposed system, we apply the LSTM model to the medical Q\&A domain. 

%% file: subfile/5_conclusion.tex
\section{Conclusions}
\label{sec:conclusions}
Nowadays, questions posted on online medical Q\&A websites are so tremendous that many of them cannot be answered in time. However, there exist many semantically similar questions in the archived Q\&A data. Thus, finding similar medical questions from the solved ones makes it possible to provide answers quickly for the unsolved questions. In this paper, we propose an end-to-end system that can find similar questions from crowdsourced medical Q\&A websites. The proposed system embeds LSTM to encode the detailed information of the unsolved medical question into a real-valued vector, and it also overcomes the issue of diversity in medical expressions. Similar questions are found based on the similarity between the vector representations of the unsolved question and the solved ones. Experimental results on the large-scale real-world dataset crawled from \textit{baobaozhidao} validate the effectiveness of the proposed system. Case studies intuitively show that the identified similar medical questions are reasonable and helpful to patients. Further, the quantitative evaluations in terms of both precision and rank correlation show that the similar questions returned by the proposed system are confirmed by human workers.